\pdfoutput=1

\documentclass{article}

\usepackage[svgnames]{xcolor}

\usepackage[preprint,nonatbib]{blueprism}

\usepackage[utf8]{inputenc} 
\usepackage[T1]{fontenc}    
\usepackage{hyperref}       
\usepackage{url}            
\usepackage{booktabs}       
\usepackage{nicefrac}       
\usepackage{microtype}      

\usepackage{amsmath}
\usepackage{amssymb}
\usepackage{amsfonts}
\usepackage{amsthm}
\usepackage{bm}
\DeclareMathOperator*{\argmax}{arg\,max}

\newcommand{\kmax}{\kappa^{(\textnormal{max})}}
\newcommand{\kmode}{\kappa^{(\textnormal{mode})}}

\usepackage[ruled,vlined,linesnumbered]{algorithm2e}
\SetKwFor{PForEach}{parallel foreach}{do}{endfch}
\SetKwFunction{isindexvalid}{is\_index\_valid}
\SetKwFunction{propagate}{propagate\_from\_neighbour}
\SetKwFunction{randomsearch}{random\_search\_around\_neighbour}

\usepackage[capitalize]{cleveref}
\usepackage{enumitem}
\usepackage{multirow}
\usepackage{graphicx}
\usepackage{siunitx}
\usepackage{subcaption}

\title{SCRAM:\ Spatially Coherent Randomized\\Attention Maps}

\author{%
    Dan A. Calian, Peter Roelants\thanks{Dan A. Calian and Peter Roelants are joint first authors with equal contribution.}\\
    \texttt{\{dan.andrei.calian,peter.roelants\}@blueprism.com}\\
    \AND
    Jacques Cali, Ben Carr, Krishna Dubba, John E. Reid, Dell Zhang\thanks{Dell Zhang is on leave from Birkbeck, University of London.}\\
    \texttt{\{jacques.cali,ben.carr,krishna.dubba,john.reid,dell.zhang\}@blueprism.com}\\
    \\\\
    Blue Prism\\
}

\begin{document}

\maketitle

\begin{abstract}
Attention mechanisms and non-local mean operations in general are key ingredients in many state-of-the-art deep learning techniques. 
In particular, the Transformer model based on multi-head self-attention has recently achieved great success in natural language processing and computer vision. 
However, the vanilla algorithm computing the Transformer of an image with $n$ pixels has $\mathcal{O}(n^2)$ complexity, which is often painfully slow and sometimes prohibitively expensive for large-scale image data. 
In this paper, we propose a fast randomized algorithm --- SCRAM --- that only requires $\mathcal{O}(n\log{n})$ time to produce an image attention map.
Such a dramatic acceleration is attributed to our insight that attention maps on real-world images usually exhibit (1) spatial coherence and (2) sparse structure. 
The central idea of SCRAM is to employ PatchMatch, a randomized correspondence algorithm, to quickly pinpoint the most compatible key (argmax) for each query first, and then exploit that knowledge to design a sparse approximation to non-local mean operations. 
Using the argmax (mode) to dynamically construct the sparse approximation distinguishes our algorithm from all of the existing sparse approximate methods and makes it very efficient. 
Moreover, SCRAM is a broadly applicable approximation to any non-local mean layer in contrast to some other sparse approximations that can only approximate self-attention.
Our preliminary experimental results suggest that SCRAM is indeed promising for speeding up or scaling up the computation of attention maps in the Transformer.
\end{abstract}

\section{Introduction}

Convolutional neural networks have long been the dominant paradigm for neural network models of images. Much of their power is attributed to their ability to efficiently calculate translation invariant local features. In contrast, attention is a mechanism by which neural networks can focus on or attend to a subset of global features and hence is capable of capturing the long range interactions necessary for many image modelling tasks.

Attention maps were initially developed to model sequences in tasks such as machine translation and were almost exclusively used in combination with recurrent networks. The sequential nature of these recurrent networks precluded their application to longer sequences and they were predominantly used in the short sequence domain. To remedy this, Vaswani et al.\ developed the Transformer attention architecture~\cite{vaswaniAttentionAllYou2017} which eschews both recurrence and convolutions. It achieved state-of-the-art performance on two WMT 2014 translation tasks. However, a quadratic run-time complexity in the length of the sequence still prevents widespread adoption of the Transformer for larger scale problems. Nevertheless, attention mechanisms continue to achieve state-of-the-art performance across text modeling tasks~\cite{daiTransformerXLAttentiveLanguage2019,luongEffectiveApproachesAttentionbased2015}, image modeling tasks~\cite{zhangSelfAttentionGenerativeAdversarial2018, yuGenerativeImageInpainting2018, belloAttentionAugmentedConvolutional2019}, video classification~\cite{wangNonlocalNeuralNetworks2018} and speech recognition~\cite{chiuMonotonicChunkwiseAttention2017}, albeit mostly on problems of limited size.

At the heart of most attention mechanisms is a per-query softmax operation over a vector of attention scores with one entry for each key to be attended to. Calculation of this softmax for every query over all keys results in a quadratic runtime complexity. Parmar et al.~\cite{parmarImageTransformer2018} and Child et al.~\cite{childGeneratingLongSequences2019} have investigated sparse approximations to this softmax. Despite empirical evidence that the structure of sparse approximations to softmax attention should depend on the data~\cite{childGeneratingLongSequences2019}, both methods use fixed structures for their sparse approximations. Our method is unique in that it uses the attention map's inputs to dynamically determine the structure of the sparse approximation. 
This is achieved using the efficient $\mathcal{O}(n\log{n})$ PatchMatch correspondence detection algorithm~\cite{barnesPatchMatchRandomizedCorrespondence2009} to suggest which keys are the best matches for every query. Thus, the structure of the sparsity in SCRAM is driven by the data. This permits greater levels of sparsity than previous approaches and consequently greater computational gains. SCRAM runs in $\mathcal{O}(n\log{n})$, a significant improvement over the $\mathcal{O}(n\sqrt{n})$ complexity of the newly emerged Sparse Transformer~\cite{childGeneratingLongSequences2019} from OpenAI (see \cref{sec:RelatedWork}).

In the rest of this paper, we first provide a precise definition of the problem that we address. 
Next, we review related work on improving the efficiency of attention maps and estimating the softmax when the number of categories, $n$, is high. 
Following that, we present our method and analyze its computational complexity. 
Finally, we present the results of preliminary experiments confirming the computational gains.

\section{Background}

Essentially, an \emph{attention} function~\cite{bahdanauNeuralMachineTranslation2014} maps a query and a set of key-value pairs to an output. 
Here the queries and keys are vectors of dimension $d_k$, while the values and outputs are vectors of dimension $d_v$. 
For each query vector $\mathbf{q}_i$, the output vector $\mathbf{o}_i$ is computed as the weighted average of the value vectors, where the weight for the $j$-th value vector $\mathbf{v}_j$, i.e., the amount of allocated attention, is given by a compatibility function of the corresponding key vector $\mathbf{k}_j$ with respect to the query vector $\mathbf{q}_i$.
In this paper, we adopt the \emph{scaled dot-product} compatibility function~\cite{vaswaniAttentionAllYou2017} that has been proven to work well in practice
\begin{equation}
    a_{ij} = \frac{\mathbf{q}_i \cdot \mathbf{k}_j}{\sqrt{d_k}} \ ,
\end{equation}
though it would be straightforward to replace the scaled dot-product with any other compatibility function. We apply a row-wise softmax to the compatibility scores, $a_{ij}$, to calculate the attention weights (or probabilities as you prefer depending on your interpretation)
\begin{equation}\label{eqn:p}
    p_{ij} = \frac{\exp(a_{ij})}{\sum_l \exp(a_{il})}\ .
\end{equation}
Thus the $i$-th output vector $\mathbf{o}_i$ is
\begin{equation}\label{eqn:softmaxoutput}
    \mathbf{o}_i
    = \mathbb{E}_{j \sim p_i}[\mathbf{v}_j]
    = \sum_{j} p_{ij} \mathbf{v}_j
    \ .
\end{equation}

This soft attention mechanism which combines a dot-product compatibility function with a softmax operation can be seen as a specialisation of \emph{non-local mean operations}~\cite{buadesNonLocalAlgorithmImage2005, wangNonlocalNeuralNetworks2018}. This framework has a more general form
\begin{equation}\label{eqn:nlmoutput}
    \mathbf{y}_i = \frac{1}{\mathcal{C}(\mathbf{q}_i, \mathbf{k})} \sum_j f(\mathbf{q}_i, \mathbf{k}_j) \mathbf{v}_j \ ,
\end{equation}
where in the soft attention specialisation we have
\begin{align}
    f(\mathbf{q}_i, \mathbf{k}_j) &= \exp \left(\frac{\mathbf{q}_i \cdot \mathbf{k}_j}{\sqrt{d_k}} \right) \ , \\
    \mathcal{C}(\mathbf{q}_i, \mathbf{k}) &= \sum_l f(\mathbf{q}_i, \mathbf{k}_j) \ .
\end{align}

\emph{Self-attention} (a.k.a.\ intra-attention) is a variant of attention in which different positions of a given input attend to different positions of the same input.  This implies that the number of queries, the number of keys, the number of values, and the number of outputs are the same. 
In the following, we denote the number of queries by $n$.
For images of height $H$ and width $W$, $n = H\times W$.

To enable the model to jointly attend to different positions from different representation sub-spaces, one can use \emph{multi-head attention}~\cite{vaswaniAttentionAllYou2017} which projects the original $d_{\text{model}}$ dimensional queries, keys and values to $d_k$, $d_k$ and $d_v$ dimensional vectors respectively via $h$ learned linear projections (resulting in $h$ heads).

The \emph{Transformer}~\cite{vaswaniAttentionAllYou2017} is a transduction model that entirely discards recurrence and convolution operations but relies solely on multi-head attention mechanism (including both self-attention and attention between input and output) to encode and decode sequences.
Due to its ability to capture long-range, subtle inter-dependencies in complex data, the Transformer has been shown to excel in a variety of deep-learning tasks~\cite{devlinBERTPretrainingDeep2018,parmarImageTransformer2018,childGeneratingLongSequences2019}. 

For a sequence of length $n$, one self-attention map in the Transformer connects each of the $n$ positions (as queries) to each of the $n$ positions (as keys) with dynamically-calculated attention weights. 
On one hand, this allows the Transformer to be much more powerful than those models with fixed, restricted connectivity patterns.
On the other hand, this requires the construction of an $n \times n$ matrix of attention weights for every attention head at every network layer. 
Obviously, such a $\mathcal{O}(n^2)$ computational complexity~\cite{vaswaniAttentionAllYou2017} makes the Transformer difficult to scale.
In practice, even using the latest GPUs, the Transformer struggles to handle large inputs with more than a few thousand elements.

\section{Related Work}
\label{sec:RelatedWork}

\subsection{Scaling up attention maps}

One well-known and widely-used engineering trick to reduce the memory usage of the Transformer is to re-compute the attention maps at ``checkpoints'' during back-propagation~\cite{chenTrainingDeepNets2016,gruslysMemoryEfficientBackpropagationTime2016,childGeneratingLongSequences2019}.
However, this does not get to the root of the problem, as the saving of memory space comes at the cost of incurring more computation (i.e., slower speed).
Furthermore, as we have mentioned above, even the computation of one single full self-attention map is intractable for very large inputs due to its quadratic computational complexity. 
Having said that, this technique is orthogonal to ours and thus can be used in conjunction.

Several works have tried to address the scalability concerns with attention maps.
These make use of several common ideas: 
sparse methods compute the attention map on a subset of the keys; 
recurrent methods propagate information from distal regions using recurrent networks; and 
summary methods calculate attention maps on summaries of sets of keys.

One examplar of sparse methods is the \emph{Image Transformer}~\cite{parmarImageTransformer2018}, a variant of the Transformer that restricts self-attention computation to a small local neighborhood of keys in the input sequence centered around the respective output position (query)~\cite{vaswaniAttentionAllYou2017} and thus increases the size of images the model can analyze.
Another is the \emph{Sparse Transformer}~\cite{childGeneratingLongSequences2019} by OpenAI\@. It is a modification of the Transformer architecture utilizing sparse factorizations to reduce the time and memory requirements from $\mathcal{O}(n^2)$ to $\mathcal{O}(n\sqrt{n})$. 
Wang et al.\ apply \emph{Non-local Neural Networks}~\cite{wangNonlocalNeuralNetworks2018} to images and videos. They use non-local mean operations~\cite{buadesNonLocalAlgorithmImage2005} (generalisations of self-attention) to model global interactions and use a fixed sparse pooling structure to improve the runtime of their method.
\emph{Monotonic Chunkwise Attention}~\cite{chiuMonotonicChunkwiseAttention2017} uses a monotonicity assumption on the ordering of the input and output to slide a soft attention window over the input.

Dai et al.\ developed \emph{Transformer-XL}~\cite{daiTransformerXLAttentiveLanguage2019}, a segment-level recurrence method with state reuse. This is a modification of the Transformer architecture that extends the context over which attention is paid whilst avoiding the quadratic complexity. Ironically they reintroduce recurrence into an architecture motivated by the desire to avoid it.

Recurrent methods that use summaries include \emph{End-to-End Memory Networks}~\cite{sukhbaatarEndToEndMemoryNetworks2015} which implements attention over a large external memory; Britz et al.\ present a method~\cite{britzEfficientAttentionUsing2017} that summarises attention scores in a fixed size memory.

\subsection{Approximating the softmax function}

Another research direction relevant to our work is the problem of approximating the softmax when the number of categories is large. This has a long history that has mostly been motivated by the size of vocabularies in NLP tasks and the consequent runtime inefficiencies when predicting words. We note that most of the approaches to this problem assume no spatial coherence in the keys and therefore struggle to achieve runtimes comparable to those possible under this assumption.

Several approaches rely on sampling~\cite{bengioAdaptiveImportanceSampling2008,gutmannNoisecontrastiveEstimationNew2010,mnihFastSimpleAlgorithm2012,zophSimpleFastNoiseContrastive2016}, while some others leverage structure in the vocabulary, to efficiently calculate the softmax~\cite{morinHierarchicalProbabilisticNeural2005,kimCharacterAwareNeuralLanguage2015}. 
We will discuss more about sampling in \cref{sec:MonteCarlo}.

\subsection{The image correspondence problem}

When the Transformer model is applied to images, the queries, keys and values correspond to the pixels (or $1 \times 1$ patches) in the given image.    
In this context, one crucial step in the computation of the Transformer model --- measuring the compatibility/similarity between each query and each key --- has actually already been studied in the fields of computer graphics and also computer vision for many years.
It is called the \emph{image correspondence} problem where the aim is to ascertain which parts of one image $\mathbf{A}$ correspond to which parts of another image $\mathbf{B}$.

PatchMatch~\cite{barnesPatchMatchRandomizedCorrespondence2009} is a highly efficient \emph{randomized} algorithm for finding approximate nearest neighbour matches between image patches. The original authors subsequently generalized PatchMatch~\cite{barnesGeneralizedPatchMatchCorrespondence2010} to handle finding multiple nearest neighbours (not just the nearest neighbour), to perform matching using arbitrary patch descriptors and distance metrics (not just on Euclidean distance between patches of RGB values), and to search across rotation and scale (not just translation). Thus, the generalized PatchMatch algorithm can be used to estimate the nearest neighbours of a set of correspondences (a.k.a.\ compatibilities).
Throughout the remaining parts of this paper, we simply refer to the \emph{generalized} version of the PatchMatch algorithm~\cite{barnesGeneralizedPatchMatchCorrespondence2010} as PatchMatch.

Specifically, given a set of queries $Q = \{\mathbf{q}_i\}$, a set of keys $K = \{\mathbf{k}_j\}$ and a function defining a correspondence between queries and keys $f: Q \times K \rightarrow \mathbb{R}$, PatchMatch estimates a set $\hat{J}_i$ of size $\kappa$ that achieves close to the maximum obtained by
\begin{equation}
    J_i = \argmax_{|J| = \kappa} \sum_{j \in J} f(\mathbf{q}_i, \mathbf{k}_j), 
    \qquad  
    \forall \ 1 \le i \le |Q|
\end{equation}

To simplify exposition we consider the case of self-attention on image data, where we define $n$ as the number of pixels in the image: $n = H \times W$; this corresponds to having $n$ keys and $n$ queries. A na\"\i ve brute force nearest neighbour search (for a single neighbour) would have a computational complexity of $\mathcal{O}(n^2)$, PatchMatch~\cite{barnesPatchMatchRandomizedCorrespondence2009} is able to achieve $\mathcal{O}(n \log{n})$ with strong convergence guarantees due to the following reasons:
\begin{itemize}[nosep,leftmargin=*]
  \item \emph{Dimensionality of output space}. It only tries to find $\mathcal{O}(n)$ matches instead of the full $\mathcal{O}(n^2)$.
  \item \emph{Natural structure of images}. Natural images exhibit spatial coherence, pixels close to each-other are likely to be similar.
  \item \emph{The law of large  numbers}. A non-trivial fraction of random nearest neighbour assignment for each query $\mathbf{q}_i \in Q$ will likely be good guesses as illustrated by Barnes et al.~\cite{barnesPatchMatchRandomizedCorrespondence2009}. 
  These good matches can be exploited to find increasingly more good matches by iteratively propagating local neighbouring matches.
\end{itemize}

When searching for multiple $\kappa$ nearest neighbours the na\"\i ve brute force method would have $\mathcal{O}(n^2 \log{\kappa})$ computational complexity, achieved using $n$ efficient $\mathcal{O}(n \log{\kappa})$ partial sorts over the correspondences of each individual query with all the keys. The serial heap variant of the multiple neighbours PatchMatch algorithm~\cite{barnesGeneralizedPatchMatchCorrespondence2010} is able to achieve a better computational complexity of $\mathcal{O}(n \log{n} \, \kappa \log{\kappa})$.

\section{Our Approach}

\subsection{Assumptions}

\begin{figure}[tb]
    \centering
    \begin{subfigure}[b]{0.49\textwidth}
        \includegraphics[width=\textwidth]{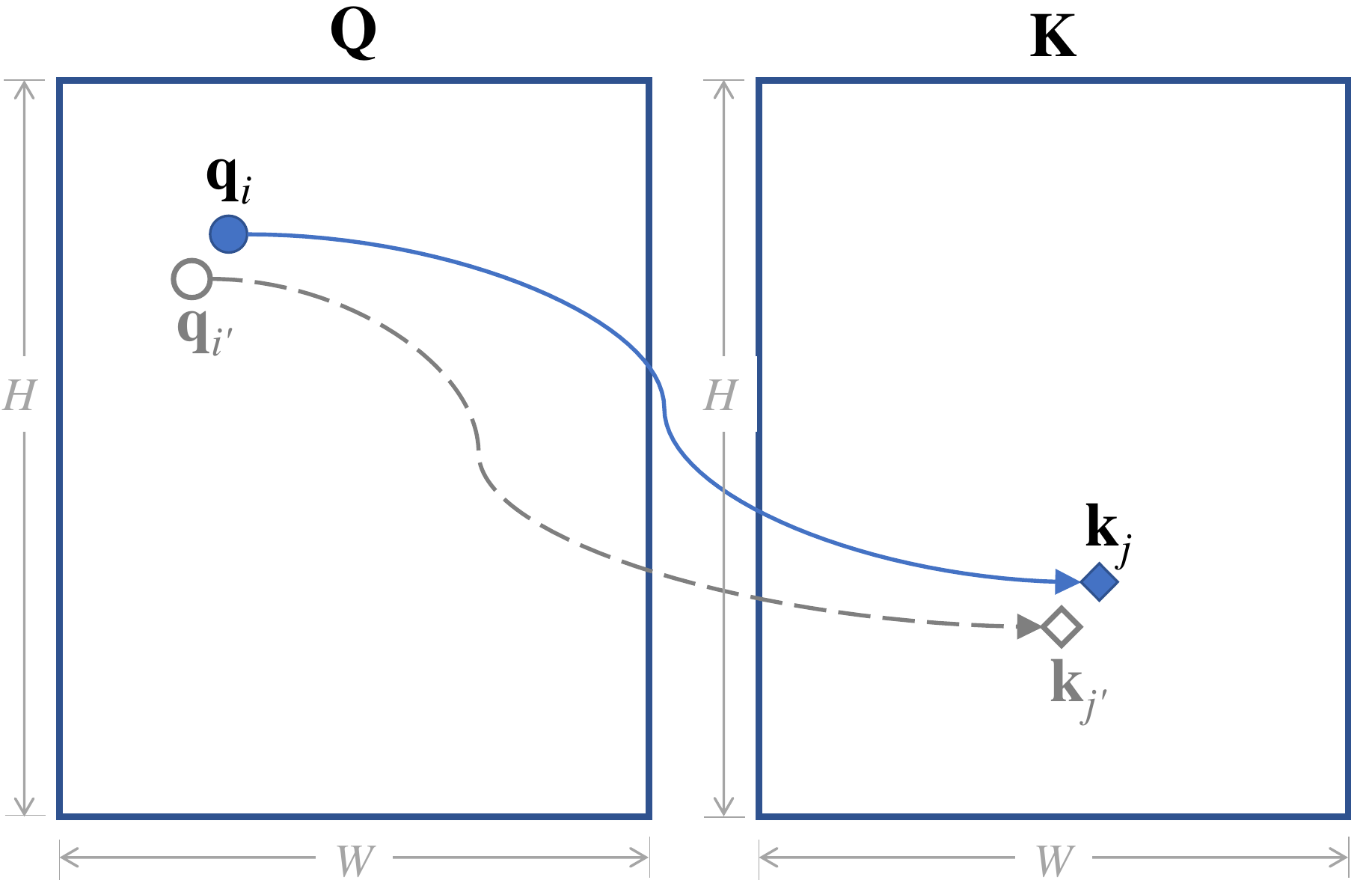}
        \caption{spatial coherence}
    \end{subfigure}
    \begin{subfigure}[b]{0.49\textwidth}
        \includegraphics[width=\textwidth]{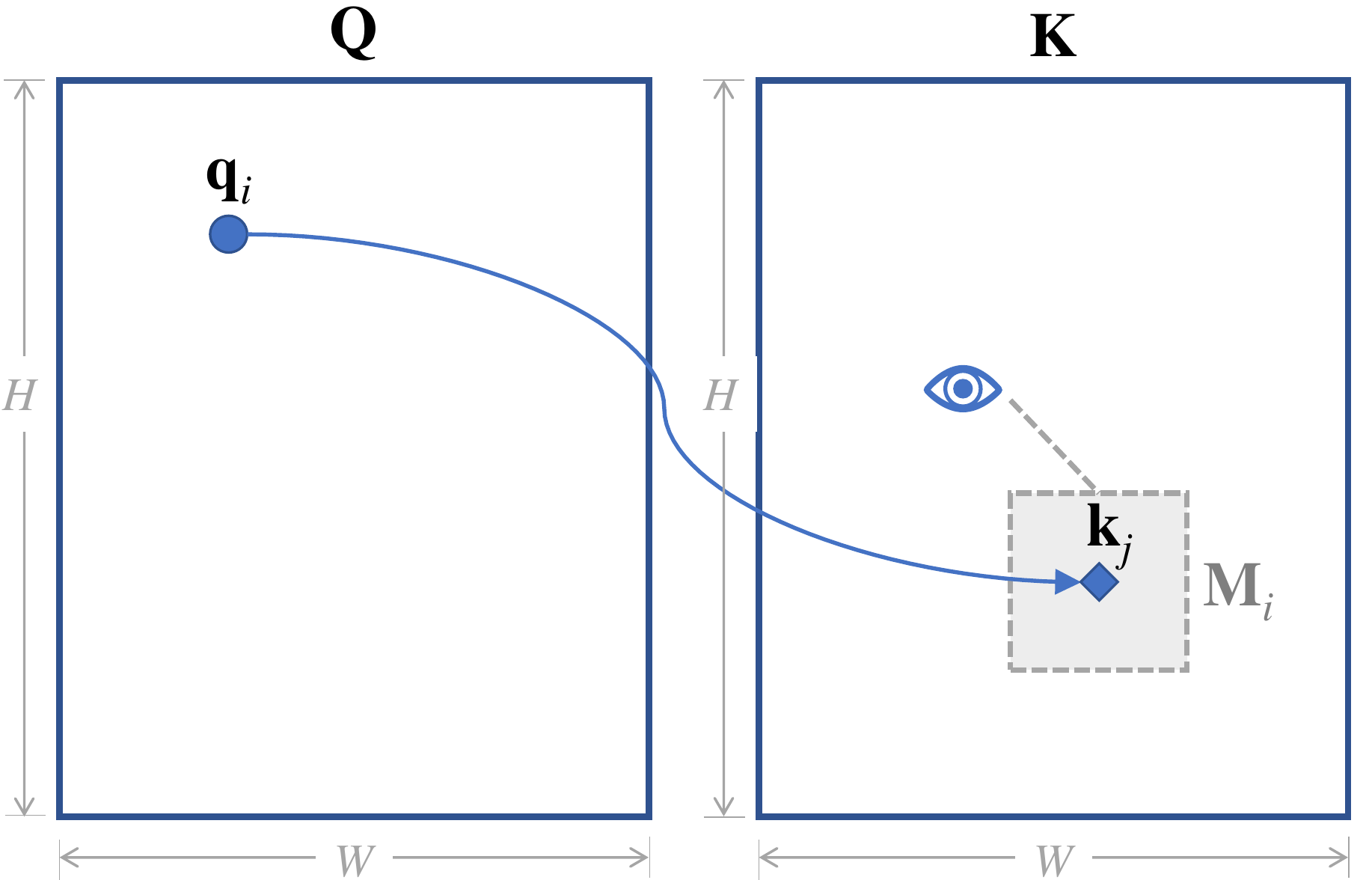}
        \caption{sparse structure}
    \end{subfigure}
    \caption{\label{fig:assumptions}To accelerate the computation of attention maps for the Transformer model, the proposed SCRAM algorithm exploits (a) the \emph{spatial coherence} of attention maps via the adapted PatchMatch algorithm and (b) the \emph{sparse structure} of attention maps via focusing on the best matching key's small local neighborhood.}
\end{figure}

Our experience suggests that visual attention maps on real-world images usually exhibit the following two properties:
\begin{enumerate}[nosep,leftmargin=*,label= (\arabic*)]
    \item \emph{spatial coherence} --- the queries close to each other tend to have similar distributions of attention over the keys, i.e., a shift in query along a spatial dimension is likely corresponding to an equal shift in key for attention;
    \item \emph{sparse structure} --- one query's attention distribution tends to concentrate on a small local area of keys, though that area is not necessarily near the respective query. 
\end{enumerate}
These observations are consistent with the previous research findings in both human vision~\cite{liaoVisualAttributeTransfer2017} and computer vision~\cite{yuGenerativeImageInpainting2018,parmarImageTransformer2018,childGeneratingLongSequences2019}.
The development of our fast approximate algorithm for attention computation is motivated by the above insights (see \cref{fig:assumptions}).

\subsection{Algorithms}

We propose to employ the PatchMatch algorithm to find the nearest neighbours field (NNF), of $\kappa$ nearest neighbours, from the queries to the keys. We use the NNF to compute a sparse approximation to the full nearest neighbour distribution (NND) (Equation~\eqref{eqn:p}) as defined by the full attention map.

Given the indices of the $\kappa$ most compatible keys $\hat{J}_i$ for query $\mathbf{q}_i$ returned from PatchMatch we define our sparse approximate distributions $\hat{p}_i$ as
\begin{equation}\label{eqn:sparse}
    \hat{p}_{ij} \propto
        \begin{cases}
            \exp\left({\mathbf{q}_i\cdot\mathbf{k}_j}/{\sqrt{d_k}}\right) & \text{if } j \in \hat{J}_i \\
            0 & \text{otherwise}
        \end{cases}
\end{equation}
and our $i$'th approximate output vector is (c.f. Equation~\eqref{eqn:softmaxoutput})
\begin{equation}
    \hat{\mathbf{o}}_i
    = \mathbb{E}_{j \sim \hat{p}_i}[\mathbf{v}_j]
    = \sum_{j \in \hat{J}_i} \hat{p}_{ij} \mathbf{v}_j
    \ .
\end{equation}

We use two distinct variants of the PatchMatch algorithm. The top-$\kmax$ variant returns an estimate of the best $\kmax$ keys for each query without regard to their spatial location. We expect the full NND to be multimodal, however the top-$\kmax$ keys will often be located proximally in the same mode. To address this, we use a top-$\kmode$ variant which finds the best $\kmode$ keys for each query whilst maintaining spatial separation as defined by a hyper-parameter $L$.

We base our PatchMatch attention algorithm on the the jump-flood GPU-version of PatchMatch~\cite{barnesPatchMatchRandomizedCorrespondence2009} which has a complexity of $\mathcal{O}(n\log{n})$ for approximating the argmax of the attention map. To compute multiple nearest keys we run the jump-flood PatchMatch argmax algorithm multiple times while enforcing: (1) non-duplicate key locations, for the top-$\kmax$ variant, and (2), spatial separation of $L$ between candidate key locations, for the top-$\kmode$ variant. We illustrate our adapted PatchMatch algorithm in Algorithm~\ref{alg:PatchMatch}; we note that the difference between top-$\kmax$ and the top-$\kmode$ variants is encapsulated in the \texttt{is\_index\_valid($\cdot$)} function, which enforces the spatial separation or non-duplication constraint on key indices.

Furthermore, as described in the main SCRAM algorithm (Alg.~\ref{alg:SCRAM}) we also expand each of the top-$\kmode$ modes into a small $(2b+1)^2$ local neighbourhood; we illustrate this process for a top-$1$ run in Figure~\ref{fig:assumptions}. The case of $b=0$ corresponds to using just the returned neighbour(s) from PatchMatch without any neighbourhood expansion. The special case of $b=0$ and $\kappa=1$ corresponds to simply computing the argmax over the keys' correspondences for each query.

The runtime complexity of our PatchMatch variants, top-$\kmax$ and top-$\kmode$, is $\mathcal{O}(n \, \kappa \, \delta \log{n})$ where $\delta$ is the runtime complexity of \texttt{is\_index\_valid($\cdot$)}, of which an optimal implementation, e.g. using a bloom filter, for avoiding invalid indices would have constant computational complexity.

\begin{algorithm}
\DontPrintSemicolon{}
    \caption{\label{alg:PatchMatch}Adapted GPU-variant of PatchMatch for 2D attention maps}
    \KwIn{
        \begin{itemize}[nosep,leftmargin=*]
            \item $\mathbf{Q}$: the ${H}\times{W}\times{d_k}$ query image whose pixel at position $(x,y)$ is represented by the query vector $\mathbf{q}_{x,y}$.
            \item $\mathbf{K}$: the ${H}\times{W}\times{d_k}$ key image whose pixel at position $(x,y)$ is represented by the key vector $\mathbf{k}_{x,y}$.
            \item $T$: the number of iterations ($T=8$ by default).
            \item $\mathbf{J}$: the $C \times H \times W \times 2$ nearest neighbour fields computed from $C$ previous runs of this algorithm.
            \item $b$: the neighbourhood size parameter.
        \end{itemize}
    }
    \smallskip
    \KwOut{
         \begin{itemize}[nosep,leftmargin=*]
            \item $\mathbf{j}$: the $H \times W \times 2$ output nearest neighbour field, where $\mathbf{j}_{x, y}$ contains the $(u, v)$ index of the best matching key $\mathbf{k}_{u,v}$ for query $\mathbf{q}_{x,y}$.
        \end{itemize}
    }
    \vspace{1em}
    \tcc{Randomly initialize the nearest neighbour field $\mathbf{j}$}
    \ForEach{position $(x,y)$}{
        Initialize $\mathbf{j}_{x, y}$ randomly with $(u,v)$ s.t. $0 \leq u < H$ and $0 \leq v < W$\;
    }
    \smallskip
    \For{$t \in [1,\ldots,T]$}{
        \PForEach{position $(x,y)$}{
            \tcc{Extract current best matching key from $\mathbf{j}$}
            $(u,v) \gets \mathbf{j}_{x, y}$\;
            \smallskip
            \tcc{Propagation step}
            \For{\texttt{jump} $\in [8, 4, 2, 1]$}{
                \For{\texttt{neighbour\_direction} $\in \{\Uparrow,\Downarrow,\Leftarrow,\Rightarrow\}$}{
                    $(r,s)\gets$ \propagate{} (\texttt{neighbour\_direction}, \texttt{jump})\;
                    \If{$\; \mathbf{q}_{x,y} \cdot \mathbf{k}_{r,s} > \mathbf{q}_{x,y} \cdot \mathbf{k}_{u,v}$ and \isindexvalid($r, s, \mathbf{J}$) \;}{
                        $(u,v) \gets (r,s)$\;
                    }
                }
            }
            \smallskip
            \tcc{Random search step}
            R = $\max(H, W)$\; 
            \For{\texttt{distance} $\in [R, \frac{R}{2}, \frac{R}{2^2}, \ldots, 1]$}{
                $(r,s )\gets$ \randomsearch{} ($u$, $v$, \texttt{distance})\;
                \If{$\; \mathbf{q}_{x,y} \cdot \mathbf{k}_{r,s} > \mathbf{q}_{x,y} \cdot \mathbf{k}_{u,v}$ and \isindexvalid($r, s, \mathbf{J}$) \;}{
                    $(u,v) \gets (r,s)$\;
                }
            }
            \smallskip
             \tcc{Update index of best matching key in $\mathbf{j}$}
            $\mathbf{j}_{x, y} \gets (u,v)$\;
        }
    }
\end{algorithm}

\begin{algorithm}
\DontPrintSemicolon{}
    \caption{\label{alg:SCRAM}SCRAM with $\kappa$ matches and $b$ local neighbourhood expansion}
    \KwIn{
        \begin{itemize}[nosep,leftmargin=*]
            \item $\mathbf{Q}$: the ${H}\times{W}\times{d_k}$ query       image whose $i$-th query       vector $\mathbf{q}_i$ represents its $i$-th pixel.
            \item $\mathbf{K}$: the ${H}\times{W}\times{d_k}$ key\:\:\:\: image whose $i$-th key\:\:\:\: vector $\mathbf{k}_i$ represents its $i$-th pixel.
            \item $\mathbf{V}$: the ${H}\times{W}\times{d_v}$ value\,     image whose $i$-th value\,     vector $\mathbf{v}_i$ represents its $i$-th pixel
            \item $b$: the neighbourhood size parameter.
        \end{itemize}
    }
    \KwOut{
        \begin{itemize}[nosep,leftmargin=*]
            \item $\widehat{\mathbf{O}}$: the ${H}\times{W}\times{d_v}$ output image whose $i$-th output vector $\hat{\mathbf{o}}_i$ represents its $i$-th pixel.
        \end{itemize}
    }
    \smallskip
    Run adapted PatchMatch several times (\cref{alg:PatchMatch}) to find the $\kappa = \kmax$, or $\kappa = \kmode$, indices of the best matching keys $\mathbf{k}_j$ for each query $\mathbf{q}_i$; with its output stored in $\hat{J}_{i}$, where $i$ corresponds to the $(x,y)$ 2D index \;
    \smallskip
    \tcc{Expand local neighbourhood around the $\kappa$ best matching keys}
    \ForEach{position $(x,y)$}{
        \For{$e \in [1,\ldots,\kappa]$}{
            Expand the ${(2b+1)}^2$ local neighbourhood centered around the $e$\textsuperscript{th} best match $(u,v)$ as \;
            $\hat{J}^{e}_{i} \gets
            \begin{bmatrix}
                (u-b,v-b) & \cdots & (u-b,v+b) \\
                \vdots    & \ddots & \vdots    \\
                (u+b,v-b) & \cdots & (u+b,v+b) \\
            \end{bmatrix}
            $\;
        }
    }
    \smallskip
    Flatten all $\kappa {(2b+1)}^2$ matches for each query $i$ into a single neighbourhood field indexed by the query $\hat{J}_{i}$ \;
    \smallskip
    \tcc{Compute output $\hat{\mathbf{o}}_i$}
    \ForEach{query $\mathbf{q}_i \in Q$}{
        Compute the sparse attention distribution (\eqref{eqn:sparse}) using the softmax function, that is\\
        $\hat{p}_{ij} =
            \begin{cases}
                \frac{
                    \exp\left({\mathbf{q}_i\cdot\mathbf{k}_j}/{\sqrt{d_k}}\right)
                }{
                    \sum_{l \in \hat{J}_{i}} \exp\left({\mathbf{q}_i\cdot\mathbf{k}_l}/{\sqrt{d_k}}\right)
                } & \text{if } j \in \hat{J}_{i} \\
                0 & \text{otherwise}
            \end{cases}
        $\;
        Compute the attention ``glimpse'' output $\hat{\mathbf{o}}_i = \mathbb{E}_{j \sim \hat{p}_i}[\mathbf{v}_j] = \sum_{j \in \hat{J}_i} \hat{p}_{ij} \mathbf{v}_j$\;
    }
\end{algorithm}

\subsubsection{Auto-regressive generative models}

To make use of this algorithm for autoregressive sequence (image) generation~\cite{vaswaniAttentionAllYou2017,parmarImageTransformer2018,childGeneratingLongSequences2019}, one just needs to mask the results of the query-key comparisons appropriately in the decoder so as to ensure that the model cannot attend to the positions that have not been generated yet. 

\subsubsection{Monte Carlo estimators}\label{sec:MonteCarlo}

Sparse approximations (Equation~\eqref{eqn:sparse}) to the sum (Equations~\eqref{eqn:softmaxoutput} or \eqref{eqn:nlmoutput}) will almost always be biased in that
\begin{equation}
    \mathbb{E}_{\hat{p}_i}[\hat{\mathbf{o}}_i] \neq \mathbf{o}_i
\end{equation}
unless for example $\mathbf{v}$ is constant over $j$. The Monte Carlo estimators we describe below do not have this drawback. Another view is to note that for our approximation to be accurate we rely on $\hat{J}_i$ covering enough of the keys with the largest attention weights. This in turn relies on our sparsity assumption. As Monte Carlo estimators use increasing numbers of samples they can overcome this dependence on the sparsity assumption as they visit more keys. On the other hand, we never wish to use $n$ or more samples as this would be as expensive as calculating the exact attention map without resorting to any approximation. We propose that Monte Carlo methods are suitable as a middle-ground between the sparse approximation and the exact calculation of the attention map. Whilst their theoretical properties in the limit of many samples are attractive we suggest they are useful as approximate methods for fewer than $n$ samples. In addition to the above considerations, the stochastic nature of Monte Carlo estimators provides an additional benefit: randomness (for example dropout) has been noted to act as a regularizer and reduce over-fitting in neural networks~\cite{srivastavaDropoutSimpleWay2014}. 
Here we describe two estimators, one based on Self-Normalized Importance Sampling and the other on the Metropolis-Hastings Markov Chain Monte Carlo (MCMC) algorithm.

\textbf{Self-Normalized Importance Sampling} (SNIS)~\cite{robertIntroducingMonteCarlo2010} is a Monte Carlo technique for estimating the expectation of a function with respect to an unnormalized probability distribution. In our method the probability distribution is defined in Equation~(\ref{eqn:p}) and the unnormalized distribution $p^{(u)}_i$ is defined by
\begin{equation}
    p^{(u)}_i(j) = p^{(u)}_{ij} = \exp(a_{ij}) = \exp\left(\frac{\mathbf{q}_i \cdot \mathbf{k}_j}{\sqrt{d_k}}\right) \ .
\end{equation}
Note that although we know the normalized distribution, the normalizing constant involves a sum over all $j$ which we are trying to avoid calculating. 
To implement SNIS, we use importance distributions $r_i(j)$ which will be defined later. 
First we describe the method. 
$T$ samples are drawn from each importance distribution
\begin{equation}
    j_{it}' \sim r_i \ , 
    \qquad
    1 \le t \le T  \ .  
\end{equation}
Each sample has an associated weight (a.k.a.\ importance or likelihood ratio)
\begin{equation}
    w_{it} = \frac{p^{(u)}_i(j_{it}')}{r_i(j_{it}')} = \frac{p^{(u)}_{ij_{jt}'}}{r_{ij_{it}'}} \ ,
\end{equation}
and the estimator that approximates $\mathbf{o}_i$ as defined in \cref{eqn:softmaxoutput} is calculated as
\begin{equation}
    \hat{\mathbf{o}}^\textnormal{IS}_i
    = \frac{\sum_t w_{it} \mathbf{v}_j}{\sum_t w_{it}}
    \approx \mathbb{E}_{j \sim p_i}[\mathbf{v}_j] \ .
\end{equation}
After calculating the estimate, diagnostics should be used to assess the performance of SNIS\@. 
An approximation to the delta method~\cite{doobLimitingDistributionsCertain1935} estimate of our estimator's variance is
\begin{equation}
    \text{Var}(\hat{\mathbf{o}}^\textnormal{IS}_{ie})
    = \sum_{t=1}^T w_{it}^2{(\mathbf{v}_{j_{it}'e} - \hat{\mathbf{o}}^\text{IS}_{ie})}^2 \ ,
    \qquad
    1 \le e \le d_v \ ,
\end{equation}
and the effective sample sizes (ESS) are
\begin{equation}
    T_i^\textnormal{ESS} = \frac{{\left(\sum_t w_{it}\right)}^2}{\sum_t w_{it}^2} 
\end{equation}
which represent how many samples we have effectively drawn from the normalized distribution $p_i$. Note that large weights reduce the ESS\@.

All that remains is to choose the importance distributions, $r_i$. These should be close to the unknown $p_i$. 
Given that we cannot obtain $r_i(j) \propto p_i(j)$ we prefer to choose $r_i$ with fatter tails than $p_i$ if possible. 
This reduces the weights $w_{it}$ obtained when rare $j_{it}'$ are sampled thus improving the ESS\@.

Given the above and making use of our knowledge of $\kmode$ modes of $p_i$ obtained from PatchMatch, we propose an importance distribution that is a mixture of local densities at each mode and a uniform distribution
\begin{equation}
    r_i(j) = r_{ij} = \frac{\alpha}{\kmode} \sum_{m=1}^{\kmode} g(j; \theta_{im}, \phi) + \frac{1 - \alpha}{n} \ ,
\end{equation}
where $g(j; \theta_{im}, \phi)$ is a local density (for example a radial basis function (RBF) based on pixel-to-pixel distances) centered at the peak of the $m$'th mode $\theta_{im}$ of $p_i$; $\phi$ is a parameter for $g$ (for example the RBF length-scale) shared across $i$ and $m$; and $\alpha$ defines the mixture proportions.

\textbf{Metropolis-Hastings} (MH)~\cite{metropolisEquationStateCalculations1953,hastingsMonteCarloSampling1970} sampling produces samples from a target distribution. These samples can be used to estimate expectations of random quantities with respect to the target distribution. In our case we have several target distributions which are again the $p_i$ as in SNIS\@. Like SNIS, MH does not need to calculate the normalization constant and will work directly with the unnormalized distribution.

We propose to run $\kmode$ chains (or a multiple thereof) for each $i$. We suggest to start each chain at one of the modes found by PatchMatch. A simple symmetric proposal distribution should be used such as a RBF based on pixel-to-pixel distances in the image.

As with SNIS, diagnostics should be used to assess the behaviour of the chains, the easiest to calculate being the acceptance rate which should not be too low. $\hat{R}$ statistics~\cite{gelmanInferenceIterativeSimulation1992} could also be applied but in practice we do not necessarily wish to generate enough samples in order to mix the chains, rather we wish each chain to explore the neighbourhood of the mode it was initialized in. For the same reason, we choose not to use a burn-in phase for the chains.

Note that SNIS is easily parallelized in both dimensions $i$ and $t$ which suits GPU implementations. In contrast, MH is inherently a sequential algorithm.
Nevertheless, opportunities for parallelism still exist: first, the estimators for those $n$ different $p_i$-s can be run in parallel; and second, all the $\kmode$ chains for each $p_i$ can be sampled concurrently.

\subsection{Discussion}

Our proposed technique falls into the category of sparse methods (see \cref{sec:RelatedWork}). 
To the best of our knowledge, Image Transformer~\cite{parmarImageTransformer2018} and Sparse Transformer~\cite{childGeneratingLongSequences2019} are the two most closely related works. 
In comparison to the former, our method also focuses attention on a small local area, but unlike their method, that small local area does not have to be around the respective query: it could be anywhere in the image, which is much more flexible and realistic.
As pointed out by Child et al.~\cite{childGeneratingLongSequences2019}, at least on some layers of the Transformer the dynamic attention could stretch over the entirety of the image, and having such a sense of the global structure would be important for autoregressive generative modeling of images and other tasks.
In comparison to the latter, our method utilizes more general global sparsity patterns of attention maps than their regular, rigid ``strided'' and ``fixed'' sparsity patterns, achieves an even lower computational complexity $\mathcal{O}(n\log{n})$, and works for not just self-attention maps but general attention maps where the set of queries is different from the set of keys.

The reason that our method can outperform those existing sparse methods is because it exploits not only the spatial coherence (of keys) \emph{within} an individual query (as existing methods do) but also the spatial coherence \emph{across} queries.
Therefore our method does have the prerequisite that the attention maps need to be computed from many queries to many keys (as in the case of self-attention in the Transformer) and those attention maps show spatial coherence collectively (as in the case of most real-world natural images).

\section{Experiments}

\begin{figure}[htb]
    \centering
    \begin{subfigure}[b]{0.49\textwidth}
        \includegraphics[width=\textwidth]{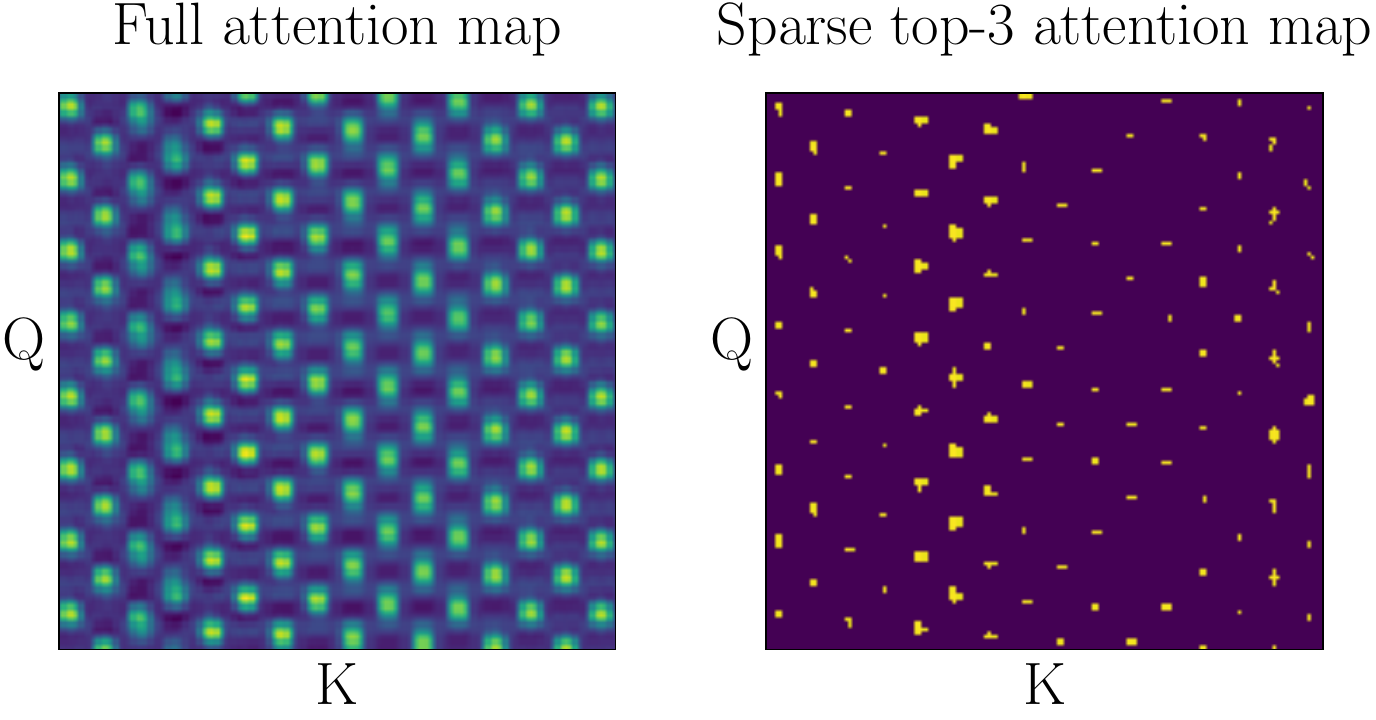}
    \end{subfigure}
    \begin{subfigure}[b]{0.49\textwidth}
        \includegraphics[width=\textwidth]{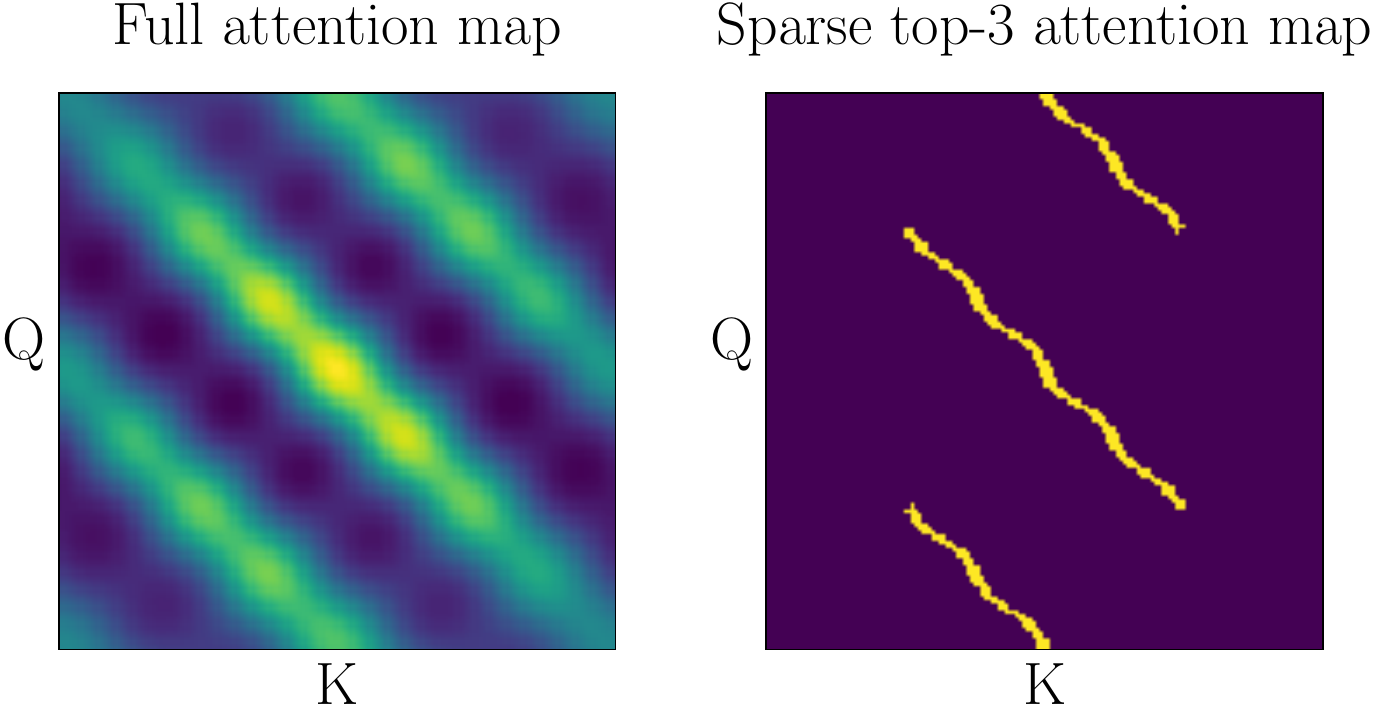}
    \end{subfigure}
    \caption{\label{fig:attn_maps_synthetic}Visualization of our top-$\kmax$ and top-$\kmode$ sparse attention maps versus the full attention map in a full dot-product attention model showing the attention map after softmax is applied per query. Note how our sparse map is consistently finding the peaks of the full attention map. The maps were generated by artificially creating queries $\mathbf{Q}$ and keys $\mathbf{K}$  as a low-rank SVD approximation ($d_k=4$) of a selected image. The $n = H \times W = 128 \times 1 = 128$ queries and keys result in the $128 \times 128$ displayed attention maps.}
\end{figure}

Visualisations of our sparse attention map between queries $\mathbf{Q}$ and keys $\mathbf{K}$ produced by top-$\kmax$ and top-$\kmode$ compared to full dot-product attention attention map as in~\cite{vaswaniAttentionAllYou2017} are shown in \cref{fig:attn_maps_synthetic}. These illustrate the sparsity of the model and the ability to consistently capture the maxima in the dot product between queries and keys.

\begin{figure}[htb]
    \centering
    \begin{subfigure}[b]{0.99\textwidth}
        \includegraphics[width=\textwidth]{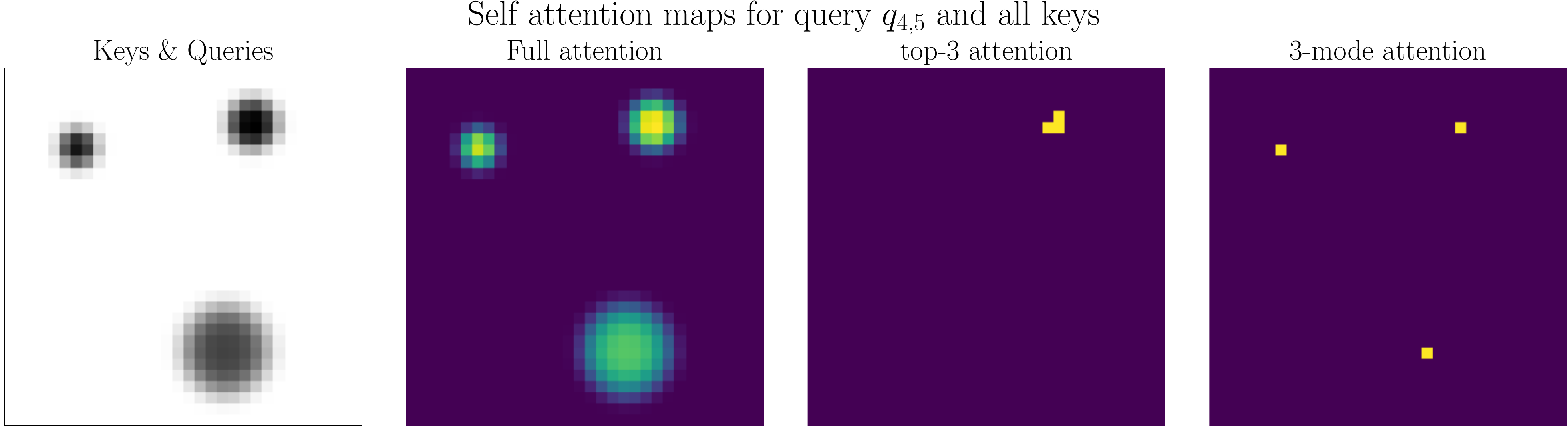}
    \end{subfigure}
    \caption{\label{fig:attn_map_mode_vs_top}Visualization of our top-$\kmax$ and top-$\kmode$ sparse attention map versus the full attention map in a full dot-product attention for self attention at query $\mathbf{q}_{4,5}$.}
\end{figure}

Comparisons of our sparse self-attention maps for a single query $\mathbf{q}$ produced by top-$\kmax$ and top-$\kmode$ are shown in \cref{fig:attn_map_mode_vs_top}. Observe how top-$\kmode$ spreads the sparse attention map over the 3 blobs in the keys $K$ while top-$\kmax$ clusters the attention around the blob with the highest (darkest) activation. 

\begin{figure}[htb]
    \centering
    \begin{subfigure}[b]{0.90\textwidth}
        \includegraphics[width=\textwidth]{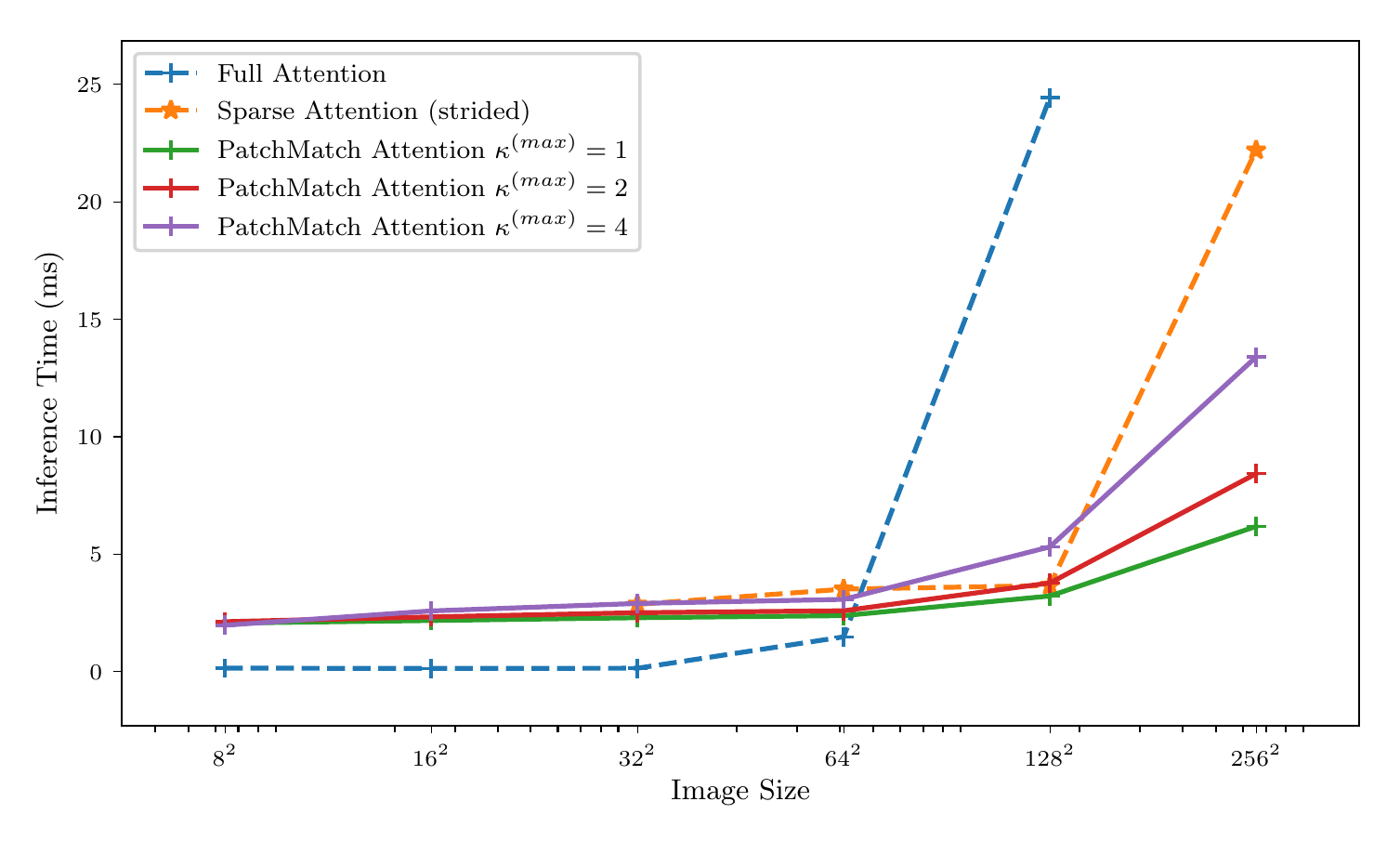}
    \end{subfigure}
    \caption{\label{fig:performance_tests}The computational cost of performing a forward pass through a single attention layer for different image sizes. Each result is an average of 5 runs. \emph{Full Attention} refers to a vanilla self-attention implementation, as used in~\cite{vaswaniAttentionAllYou2017}. \emph{Sparse Attention} refers to the sparse attention implementation from the Sparse Transformer~\cite{childGeneratingLongSequences2019}. For \emph{PatchMatch Attention}, $\kmax$ refers to the top-$\kmax$ matches. The number of feature layers $d_k$ was kept fixed at 3 with batch size $1$. Experiments were run on a single Nvidia GeForce RTX 2080 GPU.}
\end{figure}

Performance experiments were conducted comparing the computation time for the forward pass between our model, the vanilla Transformer~\cite{vaswaniAttentionAllYou2017}, and the Sparse-block Transformer~\cite{childGeneratingLongSequences2019}. Results are visualized in \cref{fig:performance_tests} demonstrating that our PatchMatch Attention attention model scales log-linear with respect to the image size. We plan to augment these graphs in the future showing memory and time usage for the backwards pass of the different models. 

In addition to the runtime performance experiments, we are working on evaluating our PatchMatch Attention module on different tasks such as image classification, detection, video recognition, and sequence generation.

\section{Conclusions}

In this paper we have proposed a novel algorithm, named \underline{S}patially \underline{C}oherent \underline{R}andomized \underline{A}ttention \underline{M}aps (SCRAM), to accelerate the computation of attention, e.g., in the Transformer, for very large images.
In comparison with the existing techniques for fast approximate attention computation, especially the recently proposed Sparse Transformer~\cite{childGeneratingLongSequences2019}, SCRAM is more general and more efficient.
As far as we know, SCRAM's $\mathcal{O}(n\log{n})$ complexity for self-attention in the Transformer is lower than that of any previous method.
 
Our main contribution is the discovery that the spatial coherence and sparse structure of attention maps could be exploited by adapting the PatchMatch algorithm to quickly and accurately guess where the ``focus'' of attention is. 
It is worth noting that such spatial coherence and sparse structure patterns can probably help improve not only the \emph{efficiency} but also the \emph{effectiveness} of attention because those properties are likely to be useful \emph{inductive biases} for the learning tasks in certain domains~\cite{childGeneratingLongSequences2019}. 
In addition, the randomization in SCRAM introduced by the usage of PatchMatch may have also played a part in the regularization of the neural network.
Further investigation into those inductive biases and regularization effects would be interesting.

Although in this paper we have focused on the fast computation of attention maps for 2D images, SCRAM is applicable to data with any number of dimensions, such as text~\cite{childGeneratingLongSequences2019}, audio~\cite{childGeneratingLongSequences2019}, and video~\cite{kalchbrennerVideoPixelNetworks2016,wangNonlocalNeuralNetworks2018}. 
In principle, there is no problem for the PatchMatch algorithm~\cite{barnesGeneralizedPatchMatchCorrespondence2010} employed by SCRAM to work on such data.
While text and audio data may contain weaker spatial coherence than image data, we expect to see stronger ``spatial'' coherence in video data due to the apparent coherence along the extra temporal dimension.
It is unclear how much impact the degree of spatial coherence has on the performance of SCRAM, and we leave an analysis for future work.

We hope that the ideas introduced in this paper are useful to the deep learning research community.
We will carry out more experiments testing SCRAM on real-world examples and release the optimized code on Github.
Stay tuned.

\small

\bibliographystyle{abbrv}
\bibliography{bp_scram}

\end{document}